\pdfoutput=1
\documentclass[11pt,a4paper]{article}
\usepackage[hyperref]{acl2020}
\usepackage{times}
\usepackage{latexsym}

\usepackage{microtype}
\usepackage{amsmath}
\usepackage{amsfonts}
\usepackage{amssymb}
\usepackage{amsthm}
\usepackage{bm}
\usepackage{multirow}
\usepackage{url}
\usepackage{array}
\usepackage{graphicx}
\usepackage{subcaption}
\usepackage{color}
\usepackage{MnSymbol}
\usepackage{makecell}
\usepackage{arydshln}
\usepackage{algorithm}
\usepackage{algorithmic}
\usepackage[shortlabels]{enumitem}
\usepackage{pifont}% http://ctan.org/pkg/pifont
\usepackage{amsmath,amsfonts,bm}

\def\eqref#1{equation~\ref{#1}}
\def\1{\bm{1}}

\DeclareMathAlphabet{\mathsfit}{\encodingdefault}{\sfdefault}{m}{sl}
\SetMathAlphabet{\mathsfit}{bold}{\encodingdefault}{\sfdefault}{bx}{n}

\usepackage{hyperref}

\usepackage{booktabs}       %
\usepackage{amsfonts}       %
\usepackage{nicefrac}       %
\usepackage{microtype}      %
\usepackage{subcaption}

\usepackage{enumitem}
\setlist{nolistsep}
\setlist[itemize]{noitemsep, topsep=0pt}

\usepackage{times}

\usepackage{graphicx} %
\usepackage{color}

\usepackage{array}
\usepackage{amssymb}
\usepackage{amsmath}
\usepackage{xspace}
\usepackage{fancyhdr}
\usepackage{comment}

\newcolumntype{H}{>{\setbox0=\hbox\bgroup}c<{\egroup}@{}}

\newcommand{\noaistats}[1]{}  %

\definecolor{darkgreen}{rgb}{0,0.4,0.0}
\definecolor{darkblue}{rgb}{0,0.1,0.3}
\definecolor{darkred}{rgb}{0.7,0.0,0.0}

\newcommand{\SUB}[1]{\ENSURE \hspace{-0.15in} \textbf{#1}}

\aclfinalcopy % Uncomment this line for the final submission
\def\aclpaperid{84} %  Enter the acl Paper ID here

\title{Towards Debiasing Sentence Representations}

\author{Paul Pu Liang, Irene Mengze Li, Emily Zheng, Yao Chong Lim,\\{\bf Ruslan Salakhutdinov, Louis-Philippe Morency}\\
Machine Learning Department and Language Technologies Institute\\
Carnegie Mellon University\\
{\tt pliang@cs.cmu.edu}
}

\date{}

\begin{document}
\maketitle

\begin{abstract}
As natural language processing methods are increasingly deployed in real-world scenarios such as healthcare, legal systems, and social science, it becomes necessary to recognize the role they potentially play in shaping social biases and stereotypes. Previous work has revealed the presence of \textit{social biases} in widely used word embeddings involving gender, race, religion, and other social constructs. While some methods were proposed to debias these word-level embeddings, there is a need to perform debiasing at the sentence-level given the recent shift towards new contextualized sentence representations such as ELMo and BERT. In this paper, we investigate the presence of social biases in sentence-level representations and propose a new method, \textsc{Sent-Debias}, to reduce these biases. We show that \textsc{Sent-Debias} is effective in removing biases, and at the same time, preserves performance on sentence-level downstream tasks such as sentiment analysis, linguistic acceptability, and natural language understanding. We hope that our work will inspire future research on characterizing and removing social biases from widely adopted sentence representations for fairer NLP.
\end{abstract}

\vspace{-1mm}
\section{Introduction}
\vspace{-1mm}

Machine learning tools for learning from language are increasingly deployed in real-world scenarios such as healthcare~\cite{VELUPILLAI201811}, legal systems~\cite{journals/nle/Dale19}, and computational social science~\cite{ws-2016-nlp-social}. Key to the success of these models are powerful \textit{embedding} layers which learn continuous representations of input information such as words, sentences, and documents from large amounts of data~\cite{DBLP:journals/corr/abs-1810-04805,mikolov2013efficient}. Although word-level embeddings~\cite{pennington2014glove,mikolov2013efficient} are highly informative features useful for a variety of tasks in Natural Language Processing (NLP), recent work has shown that word-level embeddings reflect and propagate \textit{social biases} present in training corpora~\cite{lauscher-glavas-2019-consistently,caliskan2017semantics,DBLP:journals/corr/abs-1812-08769,bolukbasi2016man}. Machine learning systems that incorporate these word embeddings can further amplify biases~\cite{sun-etal-2019-mitigating,zhao2017men,barocas2016big} and unfairly discriminate against users, particularly those from disadvantaged social groups. Fortunately, researchers working on fairness and ethics in NLP have devised methods towards \textit{debiasing} these word representations for both binary~\cite{bolukbasi2016man} and multiclass~\cite{DBLP:journals/corr/abs-1904-04047} bias attributes such as gender, race, and religion.

More recently, sentence-level representations such as ELMo~\cite{DBLP:journals/corr/abs-1802-05365}, BERT~\cite{DBLP:journals/corr/abs-1810-04805}, and GPT~\cite{radford2019language} have become the preferred choice for text sequence encoding. When compared to word-level representations, these models have achieved better performance on multiple tasks in NLP~\cite{Wu2019BetoBB}, multimodal learning~\cite{Zellers2018FromRT,Sun2019VideoBERTAJ}, and grounded language learning~\cite{Urbanek2019LearningTS}. As their usage proliferates across various real-world applications~\cite{Huang2019ClinicalBERTMC,Alsentzer2019PubliclyAC}, it becomes necessary to recognize the role they play in shaping social biases and stereotypes.

Debiasing sentence representations is difficult for two reasons. Firstly, it is usually \textit{unfeasible} to fully retrain many of the state-of-the-art sentence-based embedding models. In contrast with conventional word-level embeddings such as GloVe~\cite{pennington2014glove} and word2vec~\cite{mikolov2013efficient} which can be retrained on a single machine within a few hours, the best sentence encoders such as BERT~\cite{DBLP:journals/corr/abs-1810-04805}, and GPT~\cite{radford2019language} are trained on massive amounts of text data over hundreds of machines for several weeks. As a result, it is difficult to retrain a new sentence encoder whenever a new source of bias is uncovered from data. We therefore focus on \textit{post-hoc} debiasing techniques which add a post-training debiasing step to these sentence representations before they are used in downstream tasks~\cite{bolukbasi2016man,DBLP:journals/corr/abs-1904-04047}. Secondly, sentences display large variety in how they are composed from individual words. This variety is driven by many factors such as topics, individuals, settings, and even differences between spoken and written text. As a result, it is difficult to scale traditional word-level debiasing approaches (which involve bias-attribute words such as \textit{man, woman})~\cite{bolukbasi2016man} to sentences.

\textbf{Related Work:} Although there has been some recent work in measuring the presence of bias in sentence representations~\cite{DBLP:journals/corr/abs-1903-10561,DBLP:journals/corr/abs-1904-08783}, none of them have been able to successfully \textit{remove} bias from pretrained sentence representations. In particular,~\citet{DBLP:journals/corr/abs-1904-03310},~\citet{park-etal-2018-reducing}, and~\citet{DBLP:journals/corr/abs-1809-10610} are not able to perform post-hoc debiasing and require changing the data or underlying word embeddings and retraining which is costly.~\citet{DBLP:journals/corr/abs-1904-03035} only study word-level language models and also requires re-training. Finally,~\citet{kurita_measuring_2019} only measure bias on BERT by extending the word-level Word Embedding Association Test (WEAT)~\cite{caliskan2017semantics} metric in a manner similar to~\citet{DBLP:journals/corr/abs-1903-10561}.

In this paper, as a compelling step towards generalizing debiasing methods to sentence representations, we capture the various ways in which bias-attribute words can be used in natural sentences. This is performed by \textit{contextualizing} bias-attribute words using a \textit{diverse} set of sentence templates from various text corpora into bias-attribute sentences. We propose \textsc{Sent-Debias}, an extension of the \textsc{Hard-Debias} method~\cite{bolukbasi2016man}, to debias sentences for both binary\footnote{Although we recognize that gender is non-binary and there are many important ethical principles in the design, ascription of categories/variables to study participants, and reporting of results in studying gender as a variable in NLP~\citep{larson-2017-gender}, for the purpose of this study, we follow existing research and focus on female and male gendered terms.} and multiclass bias attributes spanning gender and religion. Key to our approach is the \textit{contextualization} step in which bias-attribute words are converted into bias-attribute sentences by using a \textit{diverse} set of sentence templates from text corpora. Our experimental results demonstrate the importance of using a large number of diverse sentence templates when estimating bias subspaces of sentence representations. Our experiments are performed on two widely popular sentence encoders BERT~\cite{DBLP:journals/corr/abs-1810-04805} and ELMo~\cite{DBLP:journals/corr/abs-1802-05365}, showing that our approach reduces the bias while preserving performance on downstream sequence tasks. We end with a discussion about possible shortcomings and present some directions for future work towards accurately characterizing and removing social biases from sentence representations for fairer NLP.

\vspace{-1mm}
\section{Debiasing Sentence Representations}
\label{debias}
\vspace{-1mm}

Our proposed method for debiasing sentence representations, \textsc{Sent-Debias}, consists of four steps: 1) \textit{defining} the words which exhibit bias attributes, 2) \textit{contextualizing} these words into bias attribute sentences and subsequently their sentence representations, 3) \textit{estimating} the sentence representation bias subspace, and finally 4) \textit{debiasing} general sentences by removing the projection onto this bias subspace. We summarize these steps in Algorithm 1 and describe the algorithmic details in the following subsections.

\begin{table}[!tbp]
    \fontsize{9}{11}\selectfont
    \begin{minipage}{.4\linewidth}
        \centering
        \setlength\tabcolsep{3.5pt}
        \begin{tabular}{c}
        \Xhline{3\arrayrulewidth}
        Binary Gender\\
        \Xhline{0.5\arrayrulewidth}
        man, woman \\
        he, she \\
        father, mother \\
        son, daughter \\
        \Xhline{3\arrayrulewidth}
        \end{tabular}
    \end{minipage}%
    \begin{minipage}{.6\linewidth}
        \centering
        \setlength\tabcolsep{3.5pt}
        \begin{tabular}{c}
        \Xhline{3\arrayrulewidth}
        Multiclass Religion\\
        \Xhline{0.5\arrayrulewidth}
        jewish, christian, muslim \\
        torah, bible, quran \\
        synagogue, church, mosque \\
        rabbi, priest, imam \\
        \Xhline{3\arrayrulewidth}
        \end{tabular}
    \end{minipage}
    \caption{Examples of word pairs to estimate the binary gender bias subspace and the $3$-class religion bias subspace in our experiments.\vspace{-4mm}}
    \label{words}
\end{table}

\newlength{\commentindent}
\setlength{\commentindent}{.3\textwidth}
\makeatletter
\renewcommand{\algorithmiccomment}[1]{\unskip\hfill\makebox[\commentindent][l]{//~#1}\par}
\LetLtxMacro{\oldalgorithmic}{\algorithmic}
\renewcommand{\algorithmic}[1][0]{%
  \oldalgorithmic[#1]%
  \renewcommand{\ALC@com}[1]{%
    \ifnum\pdfstrcmp{##1}{default}=0\else\algorithmiccomment{##1}\fi}%
}
\makeatother

\begin{figure*}
    \begin{minipage}{\linewidth}
    \begin{algorithm}[H]
    \caption{\textsc{Sent-Debias}: a debiasing algorithm for sentence representations.}
    \begin{algorithmic}[1]
        \SUB{\textsc{Sent-Debias}:}
            \STATE Initialize (usually pretrained) sentence encoder $M_\theta$.
            \STATE Define bias attributes (e.g. binary gender $g_{m}$ and $g_f$).
            \STATE Obtain words $\mathcal{D} = \{ (w_1^{(i)}, ..., w_d^{(i)}) \}_{i=1}^m$ indicative of bias attributes (e.g. Table~\ref{words}).
            \STATE $\mathcal{S} = \bigcup_{i=1}^m \textsc{Contextualize}(w_1^{(i)}, ..., w_d^{(i)}) = \{ (s_1^{(i)}, ..., s_d^{(i)}) \}_{i=1}^n$ \COMMENT{words into sentences}
            \FOR{$j \in [d]$}
                \STATE $\mathcal{R}_j = \{ M_\theta( s_j^{(i)} ) \}_{i=1}^n$ \COMMENT{get sentence representations}
            \ENDFOR
            \STATE $\mathbf{V} = \mathbf{PCA}_k \left( \bigcup_{j=1}^d \bigcup_{\mathbf{w} \in \mathcal{R}_j} \left( \mathbf{w} - \bm{\mu}_i \right) \right)$ \COMMENT{compute bias subspace}
            \FOR{each new sentence representation $\mathbf{h}$}
                \STATE $\mathbf{h}_{\mathbf{V}} = \sum_{j=1}^k \langle \mathbf{h}, \mathbf{v}_j \rangle \mathbf{v}_j$ \COMMENT{project onto bias subspace}
                \STATE $\mathbf{\hat{h}} = \mathbf{h} - \mathbf{h}_{\mathbf{V}}$ \COMMENT{subtract projection}
            \ENDFOR
    \end{algorithmic}
    \end{algorithm}
    \end{minipage}
\end{figure*}

\textbf{1) Defining Bias Attributes:} The first step involves \textit{identifying} the bias attributes and defining a set of \textit{bias attribute words} that are indicative of these attributes. For example, when characterizing bias across the male and female genders, we use the word pairs (\textit{man}, \textit{woman}), (\textit{boy}, \textit{girl}) that are indicative of gender. When estimating the $3$-class religion subspace across the Jewish, Christian, and Muslim religions, we use the tuples (\textit{Judaism}, \textit{Christianity}, \textit{Islam}), (\textit{Synagogue}, \textit{Church}, \textit{Mosque}). Each tuple should consist of words that have an equivalent meaning except for the bias attribute. In general, for $d$-class bias attributes, the set of words forms a dataset $\mathcal{D} = \{ (w_1^{(i)}, ..., w_d^{(i)}) \}_{i=1}^m$ of $m$ entries where each entry $(w_1, ..., w_d)$ is a $d$-tuple of words that are each representative of a particular bias attribute (we drop the superscript $^{(i)}$ when it is clear from the context). Table~\ref{words} shows some bias attribute words that we use to estimate the bias subspaces for binary gender and multiclass religious attributes (full pairs and triplets in appendix).

Existing methods that investigate biases tend to operate at the word-level which simplifies the problem since the set of tokens is bounded by the vocabulary size~\cite{bolukbasi2016man}. This simple approach has the advantage of identifying the presence of biases using predefined sets of word associations, and estimate the bias subspace using the predefined bias word pairs. On the other hand, the potential number of sentences are unbounded which makes it harder to precisely characterize the sentences in which bias is present or absent. Therefore, it is not trivial to directly convert these words to sentences to obtain a representation from pre-trained sentence encoders. In the subsection below, we describe our solution to this problem.

\definecolor{gg}{RGB}{15,125,15}
\definecolor{rr}{RGB}{190,45,45}

\begin{table*}[!tbp]
\fontsize{8.5}{11}\selectfont
\centering
\setlength\tabcolsep{1.5pt}
\begin{tabular}{l | c  c  c  c  c}
\Xhline{3\arrayrulewidth}
Dataset     & Type & Topics & Formality & Length & Samples \\
\Xhline{0.5\arrayrulewidth}
WikiText-2  & written & everything & formal & 24.0 & \makecell{``the mailing contained information about their history\\ and advised people to read several books,\\ which primarily focused on \{\textit{jewish}/\textit{christian}/\textit{muslim}\} history''} \\
\Xhline{0.5\arrayrulewidth}
SST	& written & movie reviews & informal & 19.2 & \makecell{``\{\textit{his}/\textit{her}\} fans walked out muttering words like horrible and terrible, \\but had so much fun dissing the film that they didn't mind the ticket cost.''} \\
\Xhline{0.5\arrayrulewidth}
Reddit & written & \makecell{politics,\\electronics,\\relationships} & informal & 13.6 & \makecell{``roommate cut my hair without my consent,\\ended up cutting \{\textit{himself}/\textit{herself}\} and is threatening to\\ call the police on me''} \\
\Xhline{0.5\arrayrulewidth}
MELD & spoken & comedy TV-series & informal & 8.1 & ``that's the kind of strength that I want in the \{\textit{man}/\textit{woman}\} I love!'' \\
\Xhline{0.5\arrayrulewidth}
POM	& spoken & opinion videos & informal & 16.0 & \makecell{``and \{\textit{his}/\textit{her}\} family is, like, incredibly confused''} \\
\Xhline{3\arrayrulewidth}
\end{tabular}
\caption{Comparison of the various datasets used to find natural sentence templates. Length represents the average length measured by the number of words in a sentence. Words in italics indicate the words used to estimating the binary gender or multiclass religion subspaces, e.g. (\textit{man}, \textit{woman}), (\textit{jewish}, \textit{christian}, \textit{muslim}). This demonstrates the variety in our naturally occurring sentence templates in terms of topics, formality, and spoken/written text.\vspace{-2mm}}
\label{datasets}
\end{table*}

\textbf{2) Contextualizing Words into Sentences:} A core step in our \textsc{Sent-Debias} approach involves \textit{contextualizing} the predefined sets of bias attribute words to sentences so that sentence encoders can be applied to obtain sentence representations. One option is to use a simple template-based design to simplify the contextual associations a sentence encoder makes with a given term, similar to how~\citet{DBLP:journals/corr/abs-1903-10561} proposed to measure (but not remove) bias in sentence representations. For example, each word can be slotted into templates such as ``This is $<$word$>$.'', ``I am a $<$word$>$.''. We take an alternative perspective and hypothesize that for a given bias attribute (e.g. gender), a single bias subspace exists across all possible sentence representations. For example, the bias subspace should be the same in the sentences ``The boy is coding.'', ``The girl is coding.'', ``The boys at the playground.'', and ``The girls at the playground.''. In order to estimate this bias subspace accurately, it becomes important to use sentence templates that are as \textit{diverse} as possible to account for all occurrences of that word in surrounding contexts. In our experiments, we empirically demonstrate that estimating the bias subspace using a large and diverse set of templates from text corpora leads to improved bias reduction as compared to using simple templates.

To capture the variety in syntax across sentences, we use large text corpora to find naturally occurring sentences. These naturally occurring sentences therefore become our sentence ``templates''. To use these templates to generate new sentences, we replace words representing a single class with another. For example, a sentence containing a male term ``\textit{he}'' is used to generate a new sentence but replacing it with the corresponding female term ``\textit{she}''. This contextualization process is repeated for all word tuples in the bias attribute word dataset $\mathcal{D}$, eventually contextualizing the given set of bias attribute words into \textit{bias attribute sentences}. Since there are multiple templates which a bias attribute word can map to, the contextualization process results in a \textit{bias attribute sentence dataset} $\mathcal{S}$ which is substantially larger in size:
\begin{align}
    \mathcal{S} &= \bigcup_{i=1}^m \textsc{Contextualize}(w_1^{(i)}, ..., w_d^{(i)})\\
    &= \{ (s_1^{(i)}, ..., s_d^{(i)}) \}_{i=1}^n, \ |\mathcal{S}| > |\mathcal{D}|
\end{align}
where $\textsc{Contextualize}(w_1, ..., w_d)$ is a function which returns a set of sentences obtained by matching words with naturally-occurring sentence templates from text corpora.

Our text corpora originate from the following five sources: 1) \textbf{WikiText-2}~\cite{DBLP:journals/corr/MerityXBS16}, a dataset of formally written Wikipedia articles (we only use the first 10\% of WikiText-2 which we found to be sufficient to capture formally written text), 2) \textbf{Stanford Sentiment Treebank}~\cite{socher-etal-2013-recursive}, a collection of 10000 polarized written movie reviews, 3) \textbf{Reddit} data collected from discussion forums related to politics, electronics, and relationships, 4) \textbf{MELD}~\citep{poria-etal-2019-meld}, a large-scale multimodal multi-party emotional dialog dataset collected from the TV-series Friends, and 5) \textbf{POM}~\cite{Park:2014:CAP:2663204.2663260}, a dataset of spoken review videos collected across 1,000 individuals spanning multiple topics. These datasets have been the subject of recent research in language understanding~\citep{DBLP:conf/iclr/MerityX0S17,DBLP:journals/corr/abs-1907-11692,DBLP:journals/corr/abs-1904-09408} and multimodal human language~\citep{liang-etal-2018-multimodal,liang-etal-2019-strong}. Table~\ref{datasets} summarizes these datasets. We also give some examples of the diverse templates that occur naturally across various individuals, settings, and in both written and spoken text.

\textbf{3) Estimating the Bias Subspace:} Now that we have contextualized all $m$ word $d$-tuples in $\mathcal{D}$ into $n$ sentence $d$-tuples $\mathcal{S}$, we pass these sentences through a pre-trained sentence encoder (e.g. BERT, ELMo) to obtain sentence representations. Suppose we have a pre-trained encoder $M_\theta$ with parameters $\theta$. Define $\mathcal{R}_j, j \in [d]$ as sets that collect all sentence representations of the $j$-th entry in the $d$-tuple, $\mathcal{R}_j = \{ M_\theta( s_j^{(i)} ) \}_{i=1}^n$. Each of these sets $\mathcal{R}_j$ defines a vector space in which a specific bias attribute is present across its contexts. For example, when dealing with binary gender bias, $\mathcal{R}_1$ (likewise $\mathcal{R}_2$) defines the space of sentence representations with a male (likewise female) context. The only difference between the representations in $\mathcal{R}_1$ versus $\mathcal{R}_2$ should be the specific bias attribute present. Define the mean of set $j$ as $\bm{\mu}_j = \frac{1}{|\mathcal{R}_j|} \sum_{\mathbf{w} \in \mathcal{R}_j} \mathbf{w}$. The bias subspace $\mathbf{V} = \{ \mathbf{v}_1, ..., \mathbf{v}_k \}$ is given by the first $k$ components of principal component analysis (PCA)~\citep{Abdi:2010:PCA:3160436.3160440}:
\begin{equation}
    \mathbf{V} = \mathbf{PCA}_k \left( \bigcup_{j=1}^d \bigcup_{\mathbf{w} \in \mathcal{R}_j} \left( \mathbf{w} - \bm{\mu}_j \right) \right).
\end{equation}
$k$ is a hyperparameter in our experiments which determines the dimension of the bias subspace. Intuitively, $\mathbf{V}$ represents the top-$k$ orthogonal directions which most represent the bias subspace.

\textbf{4) Debiasing:} Given the estimated bias subspace $\mathbf{V}$, we apply a partial version of the \textsc{Hard-Debias} algorithm~\cite{bolukbasi2016man} to remove bias from new sentence representations. Taking the example of binary gender bias, the \textsc{Hard-Debias} algorithm consists of two steps:

\textbf{4a) Neutralize:} Bias components are removed from sentences that are not gendered and should not contain gender bias (e.g., \textit{I am a doctor.}, \textit{That nurse is taking care of the patient.}) by removing the projection onto the bias subspace. More formally, given a representation $\mathbf{h}$ of a sentence and the previously estimated gender subspace $\mathbf{V} = \{ \mathbf{v}_1, ..., \mathbf{v}_k \}$, the debiased representation $\mathbf{\hat{h}}$ is given by first obtaining $\mathbf{h}_{\mathbf{V}}$, the projection of $\mathbf{h}$ onto the bias subspace $\mathbf{V}$ before subtracting $\mathbf{h}_{\mathbf{V}}$ from $\mathbf{h}$. This results in a vector that is \textit{orthogonal} to the bias subspace $\mathbf{V}$ and therefore contains no bias:
\begin{align}
    \mathbf{h}_{\mathbf{V}} &= \sum_{j=1}^k \langle \mathbf{h}, \mathbf{v}_j \rangle \mathbf{v}_j,\\
    \mathbf{\hat{h}} &= \mathbf{h} - \mathbf{h}_{\mathbf{V}}.
\end{align}

\newcolumntype{K}[1]{>{\centering\arraybackslash}p{#1}}

\begin{table*}[t]%
\fontsize{8}{11}\selectfont%
\centering
\setlength\tabcolsep{2.2pt}%
\begin{tabular}{l | *{1}{K{2.25cm}} | *{1}{K{2.25cm}} | *{1}{K{2.25cm}} | *{1}{K{2.25cm}} | *{1}{K{2.25cm}}}%
\Xhline{3\arrayrulewidth}%
Test & BERT & BERT post SST-2 & BERT post CoLA & BERT post QNLI & ELMo \\
\Xhline{0.5\arrayrulewidth}
{$\!\begin{aligned}
&\textrm{C6: M/F Names, Career/Family} \\
&\textrm{C6b: M/F Terms, Career/Family} \\
&\textrm{C7: M/F Terms, Math/Arts} \\
&\textrm{C7b: M/F Names, Math/Arts} \\
&\textrm{C8: M/F Terms, Science/Arts} \\ 
&\textrm{C8b: M/F Names, Science/Arts} \\
&\textrm{Multiclass Caliskan} \end{aligned}$} &
{$\!\begin{aligned}
+0.477 &\rightarrow \mathbf{-0.096} \\
\mathbf{+0.108} &\rightarrow -0.437 \\
+0.253 &\rightarrow \mathbf{+0.194} \\
+0.254 &\rightarrow \mathbf{+0.194} \\
+0.399 &\rightarrow \mathbf{-0.075} \\
+0.636 &\rightarrow \mathbf{+0.540} \\
+0.035 &\rightarrow \mathbf{+0.379} \end{aligned}$} &
{$\!\begin{aligned}
\mathbf{+0.036} &\rightarrow -0.109 \\
\mathbf{+0.010} &\rightarrow -0.057 \\
-0.219 &\rightarrow \mathbf{-0.221} \\
+1.153 &\rightarrow \mathbf{-0.755} \\
+0.103 &\rightarrow \mathbf{+0.081} \\
-0.222 &\rightarrow \mathbf{-0.047} \\
+1.200 &\rightarrow \mathbf{+1.000} \end{aligned}$} &
{$\!\begin{aligned}
\mathbf{-0.009} &\rightarrow +0.149 \\
+0.199 &\rightarrow \mathbf{+0.186} \\
\mathbf{+0.268} &\rightarrow +0.311 \\
\mathbf{+0.150} &\rightarrow +0.308 \\
+0.425 &\rightarrow \mathbf{-0.163} \\
\mathbf{+0.032} &\rightarrow -0.192 \\ 
+0.243 &\rightarrow \mathbf{+0.757} \end{aligned}$} &
{$\!\begin{aligned}
-0.261 &\rightarrow \mathbf{-0.054} \\
-0.155 &\rightarrow \mathbf{-0.004} \\
-0.584 &\rightarrow \mathbf{-0.083} \\
\mathbf{-0.581} &\rightarrow -0.629 \\
\mathbf{-0.087} &\rightarrow +0.716 \\
-0.521 &\rightarrow \mathbf{-0.443} \\
& \ \ -  \end{aligned}$} &
{$\!\begin{aligned}
-0.380 &\rightarrow \mathbf{-0.298} \\
-0.345 &\rightarrow \mathbf{-0.327} \\
\mathbf{-0.479} &\rightarrow -0.487 \\
+0.016 &\rightarrow \mathbf{-0.013} \\
\mathbf{-0.296} &\rightarrow -0.327 \\
+0.554 &\rightarrow \mathbf{+0.548} \\
& \ \ - \end{aligned}$} \\
\Xhline{3\arrayrulewidth}
\end{tabular}
\caption{Debiasing results on BERT and ELMo sentence representations. First six rows measure binary SEAT effect sizes for sentence-level tests, adapted from Caliskan tests. SEAT scores closer to $0$ represent lower bias. C$N$: test from~\citet{caliskan2017semantics} row $N$. The last row measures bias in a multiclass religion setting using MAC~\cite{DBLP:journals/corr/abs-1904-04047} before and after debiasing. MAC score ranges from $0$ to $2$ and closer to $1$ represents lower bias. Results are reported as $x_1 \rightarrow x_2$ where $x_1$ represents score before debiasing and $x_2$ after, with lower bias score in bold. Our method reduces bias of BERT and ELMo for the majority of binary and multiclass tests.}
%\vspace{-6mm}}
\label{debias}
\end{table*}

\textbf{4b) Equalize:} Gendered representations are centered and their bias components are equalized (e.g. \textit{man} and \textit{woman} should have bias components in opposite directions, but of the same magnitude). This ensures that any neutral words are equidistant to biased words with respect to the bias subspace. In our implementation, we skip this Equalize step because it is hard to identify all or even the majority of sentence pairs to be equalized due to the complexity of natural sentences. For example, we can never find all the sentences that \textit{man} and \textit{woman} appear in to equalize them appropriately. Note that even if the magnitudes of sentence representations are not normalized, the debiased representations are still pointing in directions orthogonal to the bias subspace. Therefore, skipping the equalize step still results in debiased sentence representations as measured by our definition of bias.

\vspace{-1mm}
\section{Experiments}
\label{exp}
\vspace{-1mm}

We test the effectiveness of \textsc{Sent-Debias} at removing biases and retaining performance on downstream tasks. All experiments are conducted on English terms and downstream tasks. We acknowledge that biases can manifest differently across different languages, in particular gendered languages~\citep{zhou-etal-2019-examining}, and emphasize the need for future extensions in these directions. Experimental details are in the appendix and code is released at \url{https://github.com/pliang279/sent_debias}.

\vspace{-1mm}
\subsection{Evaluating Biases}
\vspace{-1mm}

Biases are traditionally measured using the Word Embedding Association Test (WEAT)~\cite{caliskan2017semantics}. WEAT measures bias in word embeddings by comparing two sets of target words to two sets of attribute words. For example, to measure social bias surrounding genders with respect to careers, one could use the target words \textit{programmer, engineer, scientist}, and \textit{nurse, teacher, librarian}, and the attribute words \textit{man, male}, and \textit{woman, female}. Unbiased word representations should display no difference between the two target words in terms of their relative similarity to the two sets of attribute words. The relative similarity as measured by WEAT is commonly known as the \textit{effect size}. An effect size with absolute value closer to 0 represents lower bias.

To measure the bias present in sentence representations, we use the method as described in~\citet{DBLP:journals/corr/abs-1903-10561} which extended WEAT to the Sentence Encoder Association Test (SEAT). For a given set of words for a particular test, words are converted into sentences using a template-based method. The WEAT metric can then be applied for fixed-length, pre-trained sentence representations. To measure bias over multiple classes, we use the Mean Average Cosine similarity (MAC) metric which extends SEAT to a multiclass setting~\cite{DBLP:journals/corr/abs-1904-04047}. For the binary gender setting, we use words from the Caliskan Tests~\cite{caliskan2017semantics} which measure biases in common stereotypes surrounding gendered names with respect to careers, math, and science~\cite{articleaaaa}. To evaluate biases in the multiclass religion setting, we modify the Caliskan Tests used in~\citet{DBLP:journals/corr/abs-1903-10561} with lexicons used by~\citet{DBLP:journals/corr/abs-1904-04047}.

\vspace{-1mm}
\subsection{Debiasing Setup}
\vspace{-1mm}

We first describe the details of applying \textsc{Sent-Debias} on two widely-used sentence encoders: BERT\footnote{We used uncased BERT-Base throughout all experiments.}~\cite{DBLP:journals/corr/abs-1810-04805} and ELMo~\cite{DBLP:journals/corr/abs-1802-05365}. Note that the pre-trained BERT encoder must be fine-tuned on task-specific data. This implies that the final BERT encoder used during debiasing changes from task to task. To account for these differences, we report two sets of metrics: 1) \textbf{BERT}: simply debiasing the pre-trained BERT encoder, and 2) \textbf{BERT post task}: first fine-tuning BERT and post-processing (i.e. normalization) on a specific task before the final BERT representations are debiased. We apply \textsc{Sent-Debias} on BERT fine-tuned on two single sentence datasets, Stanford Sentiment Treebank (\textbf{SST-2}) sentiment classification~\cite{socher-etal-2013-recursive} and Corpus of Linguistic Acceptability (\textbf{CoLA}) grammatical acceptability judgment~\cite{warstadt2018neural}. It is also possible to apply BERT~\cite{DBLP:journals/corr/abs-1810-04805} on downstream tasks that involve \textit{two} sentences. The output sentence pair representation can also be debiased (after fine-tuning and normalization). We test the effect of \textsc{Sent-Debias} on Question Natural Language Inference (\textbf{QNLI})~\cite{wang-etal-2018-glue} which converts the Stanford Question Answering Dataset (SQuAD)~\cite{rajpurkar-etal-2016-squad} into a binary classification task. These results are reported as \textbf{BERT post SST-2}, \textbf{BERT post CoLA}, and \textbf{BERT post QNLI} respectively.

For ELMo, the encoder stays the same for downstream tasks (no fine-tuning on different tasks) so we just debias the ELMo sentence encoder. We report this result as \textbf{ELMo}.

\vspace{-1mm}
\subsection{Debiasing Results}
\vspace{-1mm}

We present these debiasing results in Table~\ref{debias}, and see that for both binary gender bias and multiclass religion bias, our proposed method reduces the amount of bias as measured by the given tests and metrics. The reduction in bias is most pronounced when debiasing the pre-trained BERT encoder. We also observe that simply fine-tuning the BERT encoder for specific tasks also reduces the biases present as measured by the Caliskan tests, to some extent. However, fine-tuning does not lead to consistent decreases in bias and cannot be used as a standalone debiasing method. Furthermore, fine-tuning does not give us control over which type of bias to control for and may even amplify bias if the task data is skewed towards particular biases. For example, while the bias effect size as measured by Caliskan test C7 decreases from $+0.542$ to $-0.033$ and $+0.288$ after fine-tuning on SST-2 and CoLA respectively, the effect size as measured by the multiclass Caliskan test increases from $+0.035$ to $+1.200$ and $+0.243$ after fine-tuning on SST-2 and CoLA respectively.

\begin{table}[t]
\fontsize{9}{11}\selectfont
\centering
%\fontsize{8.5}{10}\selectfont
%\begin{tabular}{l||*{5}{K{2.4cm}}}
\setlength\tabcolsep{2.5pt}
\begin{tabular}{l | c}
\Xhline{3\arrayrulewidth}
Debiasing Method & Ave. Abs. Effect Size\\
\Xhline{0.5\arrayrulewidth}
BERT original~\citep{DBLP:journals/corr/abs-1810-04805} & $+0.354$\\
FastText~\citep{bojanowski2016enriching} & $+0.565$\\
BERT word~\citep{bolukbasi2016man} & $+0.861$\\
BERT simple~\citep{DBLP:journals/corr/abs-1903-10561} & $+0.298$\\
\textsc{Sent-Debias} BERT (ours) & $\mathbf{+0.256}$\\
\Xhline{3\arrayrulewidth}
\end{tabular}
\caption{Comparison of various debiasing methods on sentence embeddings. FastText~\citep{bojanowski2016enriching} (and BERT word) derives debiased sentence embeddings with an average of debiased FastText (and BERT) word embeddings using word-level debiasing methods~\citep{bolukbasi2016man}. BERT simple adapts~\citet{DBLP:journals/corr/abs-1903-10561} by using simple templates to debias BERT representations. \textsc{Sent-Debias} BERT represents our method using diverse templates. We report the average absolute effect size across all Caliskan tests. Average scores closer to $0$ represent lower bias.\vspace{-4mm}}
\label{simple}
\end{table}

\begin{figure*}
    \centering
    \includegraphics[width=6.0cm]{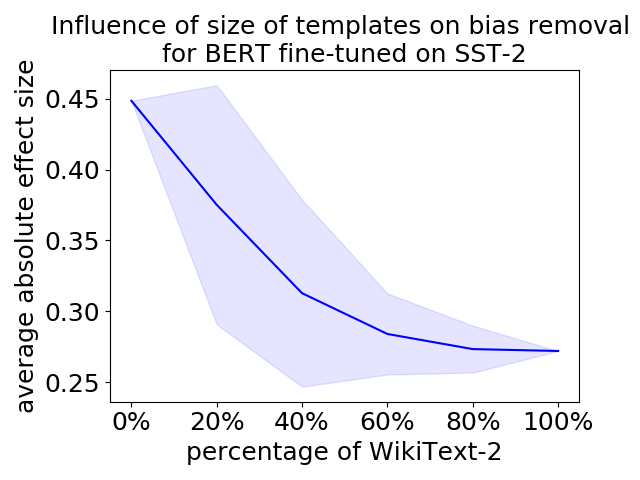}
    \includegraphics[width=6.0cm]{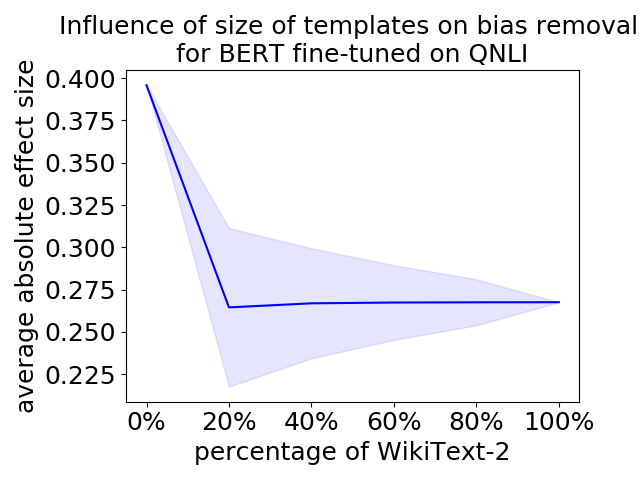}
    \caption{Influence of the number of templates on the effectiveness of bias removal on BERT fine-tuned on SST-2 (left) and BERT fine-tuned on QNLI (right). All templates are from WikiText-2. The solid line represents the mean over different combinations of domains and the shaded area represents the standard deviation. As increasing subsets of data are used, we observe a \textit{decreasing trend} and \textit{lower variance} in average absolute effect size.\vspace{-2mm}}
    \label{fig:sent_size}
\end{figure*}

\vspace{-1mm}
\subsection{Comparison with Baselines}
\vspace{-1mm}

We compare to three baseline methods for debiasing: 1) \textbf{FastText} derives debiased sentence embeddings using an average of debiased FastText word embeddings~\citep{bojanowski2016enriching} using word-level debiasing methods~\citep{bolukbasi2016man}, 2) \textbf{BERT word} obtains a debiased sentence representation from average debiased BERT word representations, again debiased using word-level debiasing methods~\citep{bolukbasi2016man}, and 3) \textbf{BERT simple} adapts~\citet{DBLP:journals/corr/abs-1903-10561} by using simple templates to debias BERT sentence representations. From Table~\ref{simple}, \textsc{Sent-Debias} achieves a lower average absolute effect size and outperforms the baselines based on debiasing at the word-level and averaging across all words. This indicates that it is not sufficient to debias words only and that biases in a sentence could arise from their debiased word constituents. In comparison with BERT simple, we observe that using diverse sentence templates obtained from naturally occurring written and spoken text makes a difference on how well we can remove biases from sentence representations. This supports our hypothesis that using increasingly diverse templates estimates a bias subspace that generalizes to different words in their context.

\vspace{-1mm}
\subsection{Effect of Templates}
\vspace{-1mm}

We further test the importance of sentence templates through two experiments.

\textbf{1) Same Domain, More Quantity:} Firstly, we ask: \textit{how does the number of sentence templates impact debiasing performance}? To answer this, we begin with the largest domain WikiText-2 (13750 templates) and divide it into 5 partitions each of size 2750. We collect sentence templates using all possible combinations of the 5 partitions and apply these sentence templates in the contextualization step of \textsc{Sent-Debias}. We then estimate the corresponding bias subspace, debias, and measure the average absolute values of all 6 SEAT effect sizes. Since different combinations of the 5 partitions result in a set of sentence templates of different sizes ($20\%$, $40\%$, $60\%$, $80\%$, $100\%$), this allows us to see the relationship between size and debiasing performance. Combinations with the same percentage of data are grouped together and for each group we compute the mean and standard deviation of the average absolute effect sizes. We perform the above steps to debias BERT fine-tuned on SST-2 and QNLI and plot these results in Figure~\ref{fig:sent_size}. Please refer to the appendix for experiments with BERT fine-tuned on CoLA, which show similar results.

For BERT fine-tuned on SST-2, we observe a \textit{decreasing trend} in the effect size as increasing subsets of the data is used. For BERT fine-tuned on QNLI, there is a decreasing trend that quickly tapers off. However, using a larger number of templates \textit{reduces the variance} in average absolute effect size and \textit{improves the stability} of the \textsc{Sent-Debias} algorithm. These observations allow us to conclude the importance of using a large \textit{number} of templates from naturally occurring text corpora.

\begin{figure*}
    \centering
    \includegraphics[width=6.0cm]{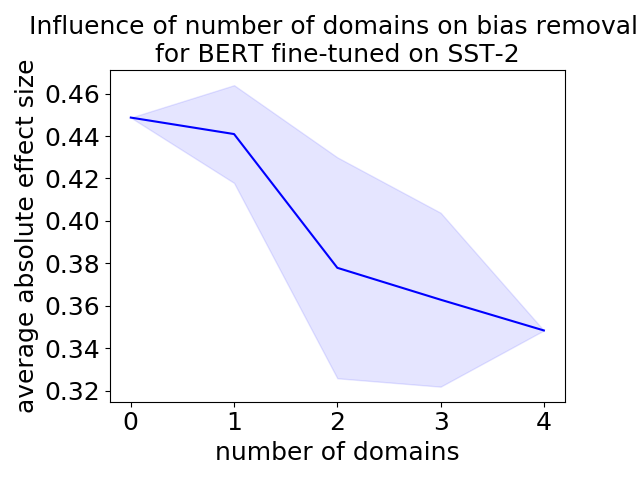}
    \includegraphics[width=6.0cm]{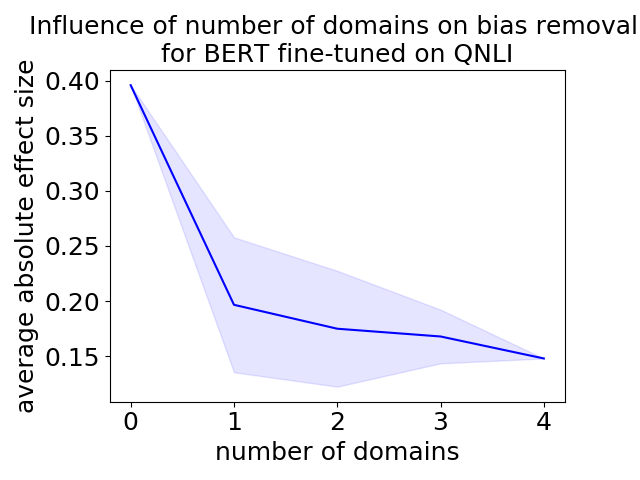}
    \caption{Influence of the number of template domains on the effectiveness of bias removal on BERT fine-tuned on SST-2 (left) and BERT fine-tuned on QNLI (right). The domains span the Reddit, SST, POM, WikiText-2 datasets. The solid line is the mean over different combinations of domains and the shaded area is the standard deviation. As more domains are used, we observe a \textit{decreasing trend} and \textit{lower variance} in average absolute effect size.\vspace{-2mm}}
    \label{fig:sent_domain}
\end{figure*}

\begin{figure*}
    \centering
    \includegraphics[width=0.8\linewidth]{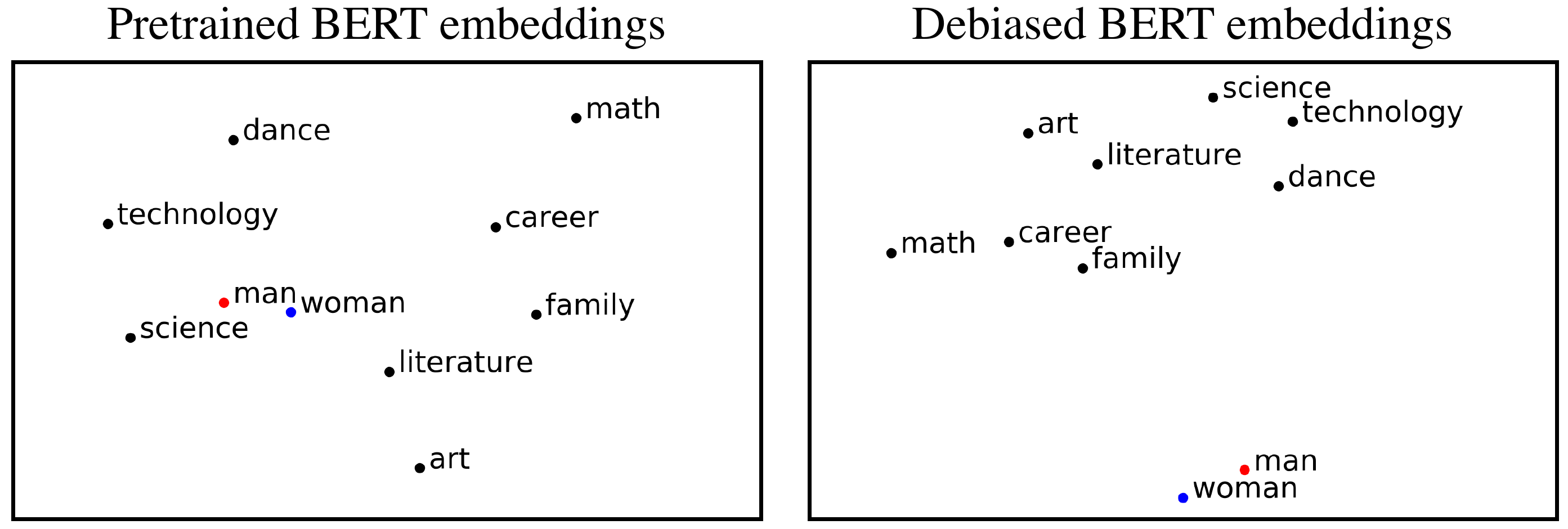}
    \caption{t-SNE plots of average sentence representations of a word across its sentence templates before (left) and after (right) debiasing. After debiasing, non gender-specific concepts (in black) are more equidistant to genders.\vspace{-2mm}}
    \label{fig:tsne}
\end{figure*}

\textbf{2) Same Quantity, More Domains:} \textit{How does the number of domains that sentence templates are extracted from impact debiasing performance?} We fix the total number of sentence templates to be $1080$ and vary the number of domains these templates are drawn from. Given a target number $k$, we first choose $k$ domains from our Reddit, SST, POM, WikiText-2 datasets and randomly sample $1080/k$ templates from each of the $k$ selected domains. We construct $1080$ templates using all possible subsets of $k$ domains and apply them in the contextualization step of \textsc{Sent-Debias}. We estimate the corresponding bias subspace, debias and measure the average absolute SEAT effect sizes. To see the relationship between the number of domains $k$ and debiasing performance, we group combinations with the same number of domains ($k$) and for each group compute the mean and standard deviation of the average absolute effect sizes. This experiment is also performed for BERT fine-tuned on SST-2 and QNLI datasets. Results are plotted in Figure~\ref{fig:sent_domain}.

We draw similar observations: there is a \textit{decreasing trend} in effect size as templates are drawn from more domains. For BERT fine-tuned on QNLI, using a larger number of domains \textit{reduces the variance} in effect size and \textit{improves stability} of the algorithm. Therefore, it is important to use a large \textit{variety} of templates across different domains.

\vspace{-1mm}
\subsection{Visualization}
\vspace{-1mm}

As a qualitative analysis of the debiasing process, we visualize how the distances between sentence representations shift after the debiasing process is performed. We average the sentence representations of a concept (e.g. \textit{man, woman, science, art}) across its contexts (sentence templates) and plot the t-SNE~\citep{vanDerMaaten2008} embeddings of these points in 2D space. From Figure~\ref{fig:tsne}, we observe that BERT average representations of \textit{science} and \textit{technology} start off closer to \textit{man} while \textit{literature} and \textit{art} are closer to \textit{woman}. After debiasing, non gender-specific concepts (e.g \textit{science, art}) become more equidistant to both \textit{man} and \textit{woman} average concepts.

\vspace{-1mm}
\subsection{Performance on Downstream Tasks}
\vspace{-1mm}

To ensure that debiasing does not hurt the performance on downstream tasks, we report the performance of our debiased BERT and ELMo on SST-2 and CoLA by training a linear classifier on top of debiased BERT sentence representations. From Table~\ref{downstream}, we observe that downstream task performance show a small decrease ranging from $1 - 3\%$ after the debiasing process. However, the performance of ELMo on SST-2 increases slightly from $89.6$ to $90.0$. We hypothesize that these differences in performance are due to the fact that CoLA tests for linguistic acceptability so it is more concerned with low-level syntactic structure such as verb usage, grammar, and tenses. As a result, changes in sentence representations across bias directions may impact its performance more. For example, sentence representations after the gender debiasing steps may display a mismatch between gendered pronouns and the sentence context. For SST, it has been shown that sentiment analysis datasets have labels that correlate with gender information and therefore contain gender bias~\citep{kiritchenko-mohammad-2018-examining}. As a result, we do expect possible decreases in accuracy after debiasing. Finally, we test the effect of \textsc{Sent-Debias} on QNLI by training a classifier on top of debiased BERT sentence pair representations. We observe little impact on task performance: our debiased BERT fine-tuned on QNLI achieves $90.6\%$ performance as compared to the $91.3\%$ we obtained without debiasing.

\vspace{-1mm}
\section{Discussion}
\label{discuss}
\vspace{-1mm}

We outline several limitations of our proposed approach as well as directions for future work

\vspace{-1mm}
\subsection{Limitations}
\vspace{-1mm}

Firstly, we would like to emphasize that both the WEAT, SEAT, and MAC metrics are not perfect since they only have positive predictive ability: they can be used to detect the presence of biases but not their absence~\citep{gonen-goldberg-2019-lipstick}. This calls for new metrics that evaluate biases and can scale to the various types of sentences appearing across different individuals, topics, and in both spoken and written text. We believe that our positive results regarding contextualizing words into sentences implies that future work can build on our algorithms and tailor them for new metrics.

\begin{table}[t]
\fontsize{9}{11}\selectfont
\centering
\setlength\tabcolsep{2.0pt}
\begin{tabular}{l | c  c | c  c}
\Xhline{3\arrayrulewidth}
\multicolumn{1}{l|}{Test} & BERT & Less biased BERT & ELMo & Less biased ELMo \\
\Xhline{0.5\arrayrulewidth}
SST-2& $92.7$ & $89.1$ & $89.6$ & $90.0$ \\
CoLA & $57.6$ & $55.4$ & $39.1$ & $37.1$ \\
QNLI & $91.3$ & $90.6$ & - & - \\
\Xhline{3\arrayrulewidth}%
\end{tabular}
\caption{We test the effect of \textsc{Sent-Debias} on both single sentence (BERT and ELMo on SST-2, CoLA) and paired sentence (BERT on QNLI) downstream tasks. The performance (higher is better) of debiased BERT and ELMo sentence representations on downstream tasks is not hurt by the debiasing step.\vspace{-4mm}}
\label{downstream}
\end{table}

Secondly, a particular bias should only be removed from words and sentences that are neutral to that attribute. For example, gender bias should not be removed from the word ``grandmother'' or the sentence ``she gave birth to me''. Previous work on debiasing word representations tackled this issue by listing all attribute specific words based on dictionary definitions and only debiasing the remaining words. However, given the complexity of natural sentences, it is extremely hard to identify the set of neutral sentences and its complement. Thus, in downstream tasks, we removed bias from all sentences which could possibly harm downstream task performance if the dataset contains a significant number of non-neutral sentences.

Thirdly, a fundamental challenge lies in the fact that these representations are trained without explicit bias control mechanisms on large amounts of naturally occurring text. Given that it becomes infeasible (in standard settings) to completely retrain these large sentence encoders for debiasing~\cite{zhao-etal-2018-learning,Zhang:2018:MUB:3278721.3278779}, future work should focus on developing better post-hoc debiasing techniques. In our experiments, we need to re-estimate the bias subspace and perform debiasing whenever the BERT encoder was fine-tuned. It remains to be seen whether there are debiasing methods which are invariant to fine-tuning, or can be efficiently re-estimated as the encoders are fine-tuned.

\vspace{-1mm}
\subsection{Ethical Considerations}
\vspace{-1mm}

Firstly, we would like to remind all users that our models, even after going through debiasing, are not perfect. They have only been examined in the definition of bias as stated in the paper and by related work~\citep{bolukbasi2016man,DBLP:journals/corr/abs-1903-10561}, and even then, does NOT remove all biases measured. In addition, there are other definitions of bias~\citep{gonen-goldberg-2019-lipstick} that researchers must also consider when applying these models to real data.

Furthermore, any model modified to have less bias can still potentially reproduce and amplify biases that appear in training data used for further fine-tuning. Therefore, we emphasize that using these model is NOT in itself enough and is not a one-size-fits-all method that automatically removes bias from sentence encoders. While we hope that some of our ideas can be useful in probing sentence encoders for bias, we strongly believe there still needs to be a critical investigation of the various ways in which bias can be introduced in each specific application these general pretrained models are used for.

\vspace{-1mm}
\section{Conclusion}
\vspace{-1mm}

This paper investigated the post-hoc removal of social biases from pretrained sentence representations. We proposed the \textsc{Sent-Debias} method that accurately captures the bias subspace of sentence representations by using a diverse set of templates from naturally occurring text corpora. Our experiments show that we can remove biases that occur in BERT and ELMo while preserving performance on downstream tasks. We also demonstrate the importance of using a large number of diverse sentence templates when estimating bias subspaces. Leveraging these developments will allow researchers to further characterize and remove social biases from sentence representations for fairer NLP.

\vspace{-1mm}
\section*{Acknowledgements}
\vspace{-1mm}

PPL and LPM were supported in part by the National Science Foundation (Awards \#1750439, \#1722822) and National Institutes of Health. RS was supported in part by US Army, ONR, Apple, and NSF IIS1763562. Any opinions, findings, and conclusions or recommendations expressed in this material are those of the author(s) and do not necessarily reflect the views of the National Science Foundation, National Institutes of Health, DARPA, and AFRL, and no official endorsement should be inferred. We also acknowledge NVIDIA's GPU support and the anonymous reviewers for their constructive comments.

\bibliography{acl2020}
\bibliographystyle{acl_natbib}

\clearpage

\appendix

\vspace{-2mm}
\section{Debiasing Details}
\vspace{-2mm}

We provide some details on estimating the bias subspaces and debiasing steps.

\textbf{Bias Attribute Words:} Table~\ref{words_supp} shows the bias attribute words we used to estimate the bias subspaces for binary gender bias and multiclass religious biases.

\textbf{Datasets:} We provide some details on dataset downloading below:

\begin{enumerate}
    \item WikiText-2 was downloaded from \url{https://github.com/pytorch/examples/tree/master/word_language_model}. We took the first 10\% of WikiText-2 sentences as naturally occurring templates representative of highly formal text.
    \item Hacker News and Reddit Subreddit data collected from news and discussion forums related to topics ranging from politics to electronics was downloaded from \url{https://github.com/minimaxir/textgenrnn/}.
\end{enumerate}

\section{Experimental Details}

\subsection{BERT}
All three variants of BERT (BERT, BERT post SST, BERT post CoLA) are uncased base model with hyper-parameters described in Table~\ref{config}.

For all three models, the second output ``pooled\_output'' of BERT is treated as the sentence embedding. The variant BERT is the pretrained model with weights downloaded from \url{https://s3.amazonaws.com/models.huggingface.co/bert/bert-base-uncased.tar.gz}. The variant BERT post SST is BERT after being fine-tuned on the Stanford Sentiment Treebank(SST-2) task, a binary single-sentence classification task \cite{socher-etal-2013-recursive}. During fine-tuning, we first normalize the sentence embedding and then feed it into a linear layer for classification. The variant BERT post CoLA is BERT fine-tuned on the Corpus of Linguistic Acceptability (CoLA) task, a binary single-sentence classification task. Normalization and classification are done exactly the same as BERT post SST. All BERT models are fine-tuned for 3 epochs which is the default hyper-parameter in the huggingface transformers repository. Debiasing for BERT models that are fine-tuned is done just before the classification layer.

\subsection{ELMo}
We use the ElmoEmbedder from allennlp.commands.elmo. We perform summation over the aggregated layer outputs. The resulting sentence representation is a time sequence vector with data dimension 1024. When computing gender direction, we perform mean pooling over the time dimension to obtain a 1024-dimensional vector for each definitional sentence. In debiasing, we remove the gender direction from each time step of each sentence representation. We then feed the debiased representation into an LSTM with hidden size 512. Finally, the last hidden state of the LSTM goes through a fully connected layer to make predictions.

\begin{table}[!tbp]
    \fontsize{8.5}{11}\selectfont
    \begin{minipage}{.5\linewidth}
        \centering
        \setlength\tabcolsep{3.5pt}
        \begin{tabular}{c}
        \Xhline{3\arrayrulewidth}
        Binary Gender\\
        \Xhline{0.5\arrayrulewidth}
        man, woman \\
        boy, girl \\
        he, she \\
        father, mother \\
        son, daughter \\
        guy, gal \\
        male, female \\
        his, her \\
        himself, herself \\
        John, Mary \\
        \Xhline{3\arrayrulewidth}
        \end{tabular}
    \end{minipage}%
    \begin{minipage}{.5\linewidth}
        \centering
        \setlength\tabcolsep{3.5pt}
        \begin{tabular}{c}
        \Xhline{3\arrayrulewidth}
        Multiclass Religion\\
        \Xhline{0.5\arrayrulewidth}
        jewish, christian, muslim \\
        jews, christians, muslims \\
        torah, bible, quran \\
        synagogue, church, mosque \\
        rabbi, priest, imam \\
        judaism, christianity, islam \\
        \Xhline{3\arrayrulewidth}
        \end{tabular}
    \end{minipage}
    \caption{Word pairs to estimate the binary gender bias subspace (left) and the $3$-class religion bias subspace (right).\vspace{-2mm}}
    \label{words_supp}
\end{table}

%\newcolumntype{K}[1]{>{\centering\arraybackslash}p{#1}}

% \setlength{\tabcolsep}{2pt}
\begin{table}[t]
\fontsize{8.5}{11}\selectfont
\centering
%\fontsize{8.5}{10}\selectfont
%\begin{tabular}{l||*{5}{K{2.4cm}}}
\setlength\tabcolsep{3.5pt}
\begin{tabular}{l | l}
\Xhline{3\arrayrulewidth}
Hyper-parameter & Value \\
\Xhline{0.5\arrayrulewidth}
{$\!\begin{aligned}
&\textrm{attention\_probs\_dropout\_prob} \\
&\textrm{hidden\_act} \\
&\textrm{hidden\_dropout\_prob} \\
&\textrm{hidden\_size} \\
&\textrm{initializer\_range} \\ 
&\textrm{intermediate\_size} \\
&\textrm{max\_position\_embeddings} \\
&\textrm{num\_attention\_heads} \\
&\textrm{num\_hidden\_layers} \\
&\textrm{type\_vocab\_size} \\
&\textrm{vocab\_size}
\end{aligned}$} &
{$\!\begin{aligned}
& 0.1 \\
& \textrm{gelu} \\
& 0.1 \\
& 768 \\
& 0.02 \\
& 3072 \\
& 512 \\
& 12 \\
& 12 \\
& 2 \\
& 30522 \end{aligned}$} \\
\Xhline{3\arrayrulewidth}
\end{tabular}
\caption{Configuration of BERT models, including BERT, BERT$\rightarrow$ SST, and BERT$\rightarrow$ CoLA.}
\label{config}
\end{table}

\begin{figure*}[t]
    \centering
    \includegraphics[width=7.0cm]{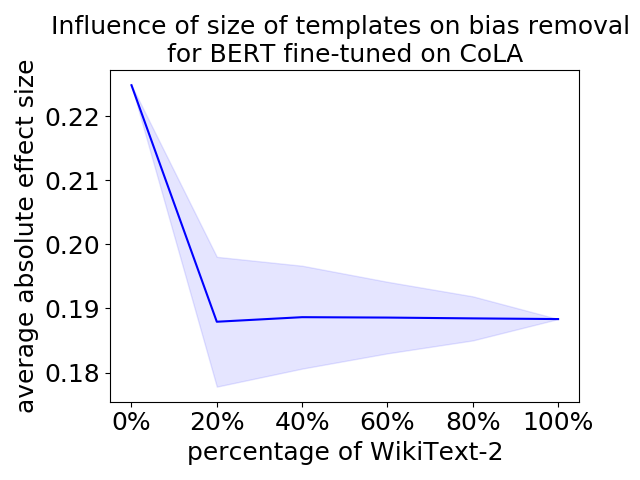}
    \includegraphics[width=7.0cm]{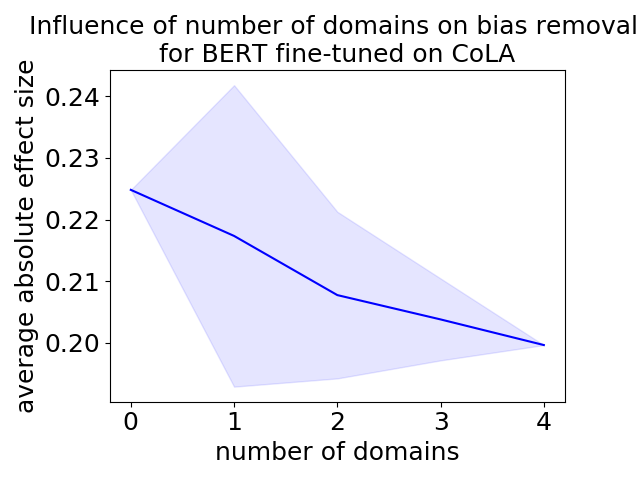}
    \caption{Evaluation of Bias Removal on BERT fine-tuned on CoLA with varying percentage of data from a single domain (left) and varying number of domains with fixed total size (right). }
    \label{fig:cola_diversity}
\end{figure*}

\vspace{-2mm}
\section{Additional Results}
\vspace{-2mm}

We also studied the effect of templates on BERT fine-tuned on CoLA as well. Steps taken are exactly the same as described in \textbf{Effect of Templates: Same Domain, More Quantity} and \textbf{Effect of Templates: Same Quantity, More Domains}. Results are plotted in Figure~\ref{fig:cola_diversity}. It shows that debiasing performance improves and stabilizes with the number of sentence templates as well as the number of domains.

\end{document}